\documentclass[journal]{IEEEtran}
\usepackage[numbers,sort&compress]{natbib}
\ifCLASSINFOpdf
   \usepackage[pdftex]{graphicx}
   \graphicspath{{../pdf/}{../jpeg/}}
   \DeclareGraphicsExtensions{.pdf,.jpeg,.png}
\else
   \usepackage[dvips]{graphicx}
   \graphicspath{{../eps/}}
   \DeclareGraphicsExtensions{.eps}
\fi
\usepackage{amsmath}
\usepackage{multirow}
\ifCLASSOPTIONcompsoc
  \usepackage[caption=false,font=normalsize,labelfont=sf,textfont=sf]{subfig}
\else
  \usepackage[caption=false,font=footnotesize]{subfig}
\fi
\usepackage{url}

\begin{document}
%
\title{Aesthetic Attributes Assessment of Images with AMANv2 and DPC-CaptionsV2}

%
%
%

\author{Xinghui Zhou, Xin Jin*, Jianwen Lv, Heng Huang, Ming Mao, Shuai Cui
\thanks{Xinghui Zhou is with the School of Cyber Science and Technology, University of Science and Technology of China, Hefei, China. E-mail: zhouxinghui@ustc.edu.cn.}
\thanks{Xin Jin is with the Department of Cyber Security, Beijing Electronic Science and Technology Institute, Beijing, China, and also with the Beijing Institute for General Artificial Intelligence (BIGAI), Beijing, China. Xin Jin* is the corresponding author. E-mail: jinxin@besti.edu.cn.}
\thanks{Jianwen Lv, Heng Huang and Ming Mao are with the Department of Cyber Security, Beijing Electronic Science and Technology Institute, Beijing, China. }
\thanks{Shuai Cui is with the Department of Philosophy and Mathematics, University of California, Davis, CA, USA.}
}

%
%

\markboth{IEEE Transactions on Multimedia,~Vol.~XX, No.~X, October~2021}%
{Shell \MakeLowercase{\textit{et al.}}: Bare Demo of IEEEtran.cls for IEEE Journals}
%



\maketitle

\begin{figure*}
	\centering
	\includegraphics[width=\textwidth]{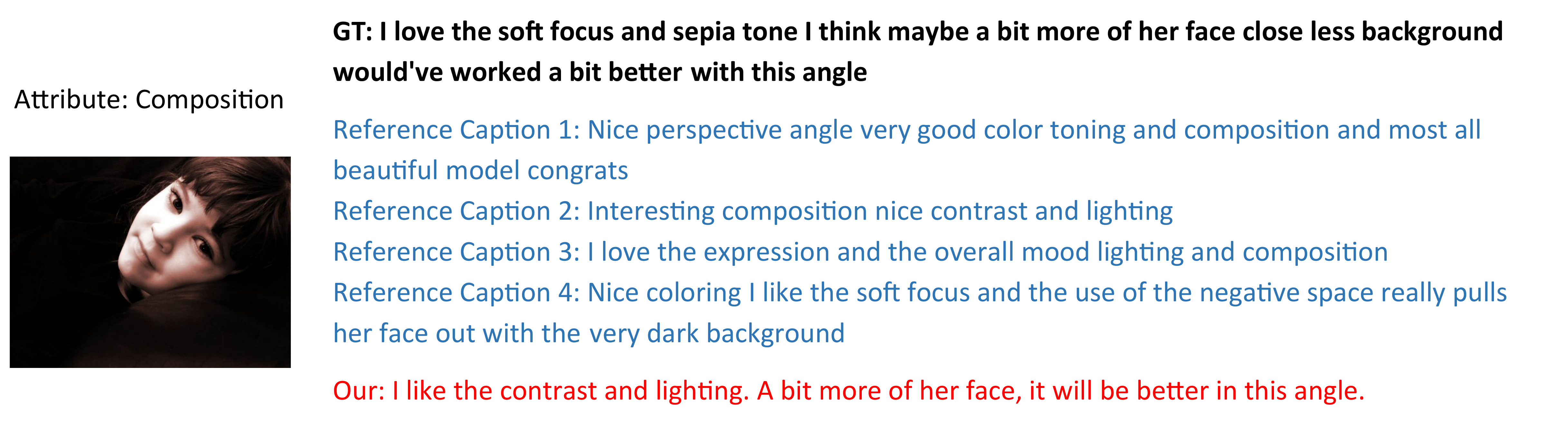}
	\caption{Aesthetic Attributes Assessment of Images. In our new DPC-CaptionsV2, we choose 5 captions for each aesthetic attribute of an image (composition in this figure). The one with the largest weight of the bag of words model is considered as the ground truth value. The others are used to avoid wrong options and increase the relevance of generated tokens.}
	\label{fig:teaser}
\end{figure*}

\begin{abstract}
Image aesthetic quality assessment is popular during the last decade. Besides numerical assessment, nature language assessment (aesthetic captioning) has been proposed to describe the generally aesthetic impression of an image. In this paper, we propose aesthetic attribute assessment, which is the aesthetic attributes captioning, i.e., to assess the aesthetic attributes such as composition, lighting usage and color arrangement. 
It is a non-trivial task to label the comments of aesthetic attributes, which limit the scale of the corresponding datasets. We construct a novel dataset, named \emph{DPC-CaptionsV2}, by a semi-automatic way. The knowledge is transferred from a small-scale dataset with full annotations to large-scale professional comments from a photography website. Images of DPC-CaptionsV2 contain comments up to 4 aesthetic attributes: composition, lighting, color, and subject. Then, we propose a new version of Aesthetic Multi-Attributes Networks (AMANv2) based on the BUTD model and the VLPSA model. AMANv2 fuses features of a mixture of small-scale PCCD dataset with full annotations and large-scale DPCCaptionsV2 dataset with full annotations. The experimental results of DPCCaptionsV2 show that our method can predict the comments on 4 aesthetic attributes, which are closer to aesthetic topics than those produced by the previous AMAN model. Through the evaluation criteria of image captioning, the specially designed AMANv2 model is better to the CNN-LSTM model and the AMAN model.
\end{abstract}


%
\IEEEpeerreviewmaketitle

\section{Introduction}
%
%
%
%
\IEEEPARstart{I}{mage} Aesthetic Quality Assessment (IAQA) is the aesthetic assessment of images. In the last decades, IAQA is very popular in the fields of computer vision, computational aesthetics, psychology and neuroscience. Aesthetic quality assessment also provides important reference for image quality assessment (IQA) in some scenarios (such as photography and post-editing). 

There are two categories in most researches: high aesthetic quality (professional) and low aesthetic quality (amateur). The second popular assessment task is to give a continuously numerical score of aesthetics. Besides, numerical assessment task is to predict a score distribution of human rating in aesthetics is \cite{IEEEhowto:CuiTMM2018,IEEEhowto:JinAAAI2018,IEEEhowto:NIMATIP18}.

For a human artist, when shown a photo or a drawing, he/she will not just give a numerical score but always say a paragraph to describe many aesthetic attributes such as composition, lighting, color, focus of the image. In the image, not only the objects and the edge need to be detected, higher-dimensional image features, such as the proportion of the object, the collocations with fashion and the camera techniques, are more important manifestations, and captions can show them well. Zhou et al. \cite{IEEEhowto:ZhouACMMM16} construct the AVA-Comments dataset with only visual features. They improve the performance of aesthetics assessment on the AVA dataset but giving comments for an image. Pioneer work of Chang et al. \cite{IEEEhowto:ChangICCV17} proposes aesthetic captioning of images. They build Photo Critique Captioning Dataset (PCCD) for the community. The PCCD contains 4,235 images and 29,645 comments. Each image is attached with comments and scores of 7 aesthetic attributes. However, they only output a sentence of assessment, which can not give a full review of aesthetic attributes. The value of PCCD is not fully explored. Besides, the size of PCCD is relatively small compared to AVA dataset \cite{IEEEhowto:MurrayCVPR2012}, which is commonly used in this field but do not contain ground truth of aesthetic captions and attributes. Jin et al. \cite{IEEEhowto:Jin_19_ACMMM} proposed the DPC-Caption data set, and for the first time combined image captions of multiple attributes with the task of aesthetic quality evaluation.

In this work, we propose \emph{Aesthetic Attributes Assessment of Images}, as shown in Figure \ref{fig:teaser}. We predict aesthetic attributes captions and the aesthetic score of each attribute. Based on the previous work, we build a better dataset, named \emph{DPC-CaptionsV2} by the aesthetic knowledge transferring method from dpchallenge.com. DPC-CaptionsV2 contains comments up to 5 aesthetic attributes of one image. There are 112,000 images and 560,000 comments. Then, we propose aesthetic multi-attribute network version 2, which contains a base image captioning network (such as bottom-up and top-down attention network and object-semantics transformer layers) and aesthetic features fusion network. We train this model on both small-scale PCCD dataset (4,235 images and 29,645 comments), containing attribute comments and scores, and our large-scale DPC-CaptionsV2 dataset with only attribute comments. We evaluate our method of captioning of attributes on DPC-CaptionsV2 with image captioning criteria.

As we know, this is the first work that can produce captions for aesthetic attributes of an image. In the early version of this paper \cite{IEEEhowto:Jin_19_ACMMM}, we proposed method Aesthetic Multi-Attribute Network (AMAN) and dataset DPC-Captions. Constructing the dataset depends on the proposed method: transferring from a small-scale image dataset with full annotations to the large-scale dataset with weak annotations. We extend the previous work badly, and the main differences include:

\begin{itemize}
	\item In this paper we propose DPC-CaptionsV2 (92,006 images and 392,625 comments). The aesthetic attributes are classified into four main attributes of photography: composition, lighting, color, subject. We update the comment matching algorithm from simple keyword matching in DPC-Captions to Bag of Words (BoW) with a BERT-based text classification model. That reduces the bias of BoW to aesthetic comments in PCCD. Using the new matching algorithm, for each attribute, there are up to 5 comments that can be matched. We choose one of them as the ground truth caption and the others as reference ones, which make the output captions more stable. Compared with DPC-Captions, our new version provides more extensive aesthetic captions, which greatly improves the quality of the dataset.
	
	\item We propose Aesthetic Multi-Attribute Network version 2 (AMANv2), which uses a two-stage training processes on a small-scale full annotated dataset and a large-scale weakly annotated one. Smaller datasets are used to filter the required aesthetic comments from the original data. On the larger DPC-CaptionV2 dataset, we improve Bottom-Up and Top-Down Attention (BUTD) and Visual Language Pre-training and Self-Attention (VLPSA), and adds small-scale, based embedding, layers to the model. The aesthetic attributes of the data set can predict the model regression on the required target. Based on aesthetic multi-attributes and the above attention networks, we propose Aesthetic Multi-Attribute Network version 2 (AMANv2).
\end{itemize}

\section{Related Work}
\textbf{Image Aesthetic Quality Assessment}. Before deep learning era, many hand-crafted features \cite{IEEEhowto:ChenTIP2015,IEEEhowto:JinECCV2010} are designed for aesthetic image classification and scoring as surveyed by Deng et al. \cite{IEEEhowto: DengSPM2017}. Deep learning methods are proposed recently for aesthetic assessment\cite{IEEEhowto: DongNC2015,IEEEhowto:JinWCSP2016,IEEEhowto:JinAAAI2018,IEEEhowto:KaoArXiv2016,IEEEhowto:KaoSPIC2016,IEEEhowto:KongECCV2016,IEEEhowto:LuMM2014,IEEEhowto:MaCVPR2017,IEEEhowto:MaiCVPR2016,IEEEhowto:WangSP2016}.
They outperform traditional methods. Lu et al. \cite{IEEEhowto:LuMM2014} propose a two column CNNs for binary classification. Mai et al. \cite{IEEEhowto:MaiCVPR2016} introduce ratio-preserving assessment of aesthetics by using SPP. Kong et al. \cite{IEEEhowto:KongECCV2016} propose the AADB dataset which contains scores of 12 aesthetic attributes and use a rank-preserving loss for aesthetic scoring. Kao et al. \cite{IEEEhowto:KaoArXiv2016} propose an aesthetics assessment combing the semantic classification of a image. Jin et al. \cite{IEEEhowto:JinAAAI2018} propose CJS-CNN for aesthetic score distribution prediction. In the above literatures, they only consider numerical assessment without taking the aesthetic assessment by languages into consideration.
Other methods are also used for image aesthetic evaluation, such as Pfister et al. \cite{IEEEhowto:Pfister2021} uses self-supervised learning to replace common ImageNet-based pre-training models. Regarding the composition's info of the image, Liu et al. \cite{IEEEhowto:Liu2019} propose a full convolutional network as the feature encoder of the input image, and uses the encoded feature map to represent the composition information in the image, and further graph convolution infers the image. Zeng et al. \cite{IEEEhowto:Zeng2019} proposed a unified probability formula for three image aesthetics evaluation tasks (binary classification, average score regression and score distribution prediction), and improved the noisy original score distribution to obtain a data distribution with stronger generalization ability. Kong et al. \cite{IEEEhowto:KuangTMM2020} proposed a method of UAV video quality assessment.

\textbf{New tasks and datasets of IAQA}. Kong et al. \cite{IEEEhowto:KongECCV2016} design a dataset named AADB (Aesthetics and Attributes Database) which contains 8 aesthetic attributes of each image. However, the label of each aesthetic attribute is only binary value (\emph{good} or \emph{bad}). Zhou et al. \cite{IEEEhowto:AVAComments} design a dataset named AVA-Comments, which adds comments from DPChallenge.com to AVA dataset \cite{IEEEhowto:MurrayCVPR2012} which only contains aesthetic score distributions of images. Zhou et al. use the image and the attached comments to give a binary classification of aesthetics. Wang et al. \cite{IEEEhowto:AVAReviews} design a dataset named AVA-Reviews, which selects 40,000 images from AVA dataset and contains 240,000 reviews. Chang et al. \cite{IEEEhowto:ChangICCV17} design PCCD dataset, which contains 4,235 images and 29,645 comments. However, both \cite{IEEEhowto:AVAReviews} and \cite{IEEEhowto:ChangICCV17} can only give a single sentence as the comments of the aesthetics of an image. They do not describe the individual aesthetic attributes. \cite{IEEEhowto:Ghosal2019} proposed a probabilistic aesthetic caption-filtering method for cleaning internet data to generate a larger dataset: AVACaptions,which has more higher quality captions than PCCD. \cite{IEEEhowto:Zou2020} proposed the task of food image aesthetic captioning and decomposed it into two tasks: single-aspect captioning and an unsupervised text compression. They also collect a dataset which contains comments related to six aesthetic attributes.

\textbf{Image Captioning.} Most work of image captioning follow CNN-RNN framework and achieve great results \cite{IEEEhowto: DonahueHRVGSD17,IEEEhowto:KarpathyF17,IEEEhowto:MaoHTCY016}.
Most of recent literatures of image captioning \cite{IEEEhowto:Chen_2018_CVPR,IEEEhowto:Aneja_2018_CVPR,IEEEhowto:Anderson_2018_CVPR,IEEEhowto:Luo_2018_CVPR,IEEEhowto:Mathews_2018_CVPR}  introduce attention scheme. We follow this trend and add attention model in our network.

Recent studies \cite{IEEEhowto:Lu2019,IEEEhowto:Tan2019,IEEEhowto:Chen2020,IEEEhowto:Su2019} on vision-language pretraining (VLP) have shown that it can effectively learn generic representations from massive image-text pairs, and that fine-tuning VLP models on task-specific data achieves state-of-the-art (SoTA) results on well-established visual and language tasks. The latest research results are OSCAR (\textbf{O}bject-\textbf{S}emanti\textbf{c}s \textbf{A}ligned P\textbf{r}e-training)\cite{IEEEhowto:Li2020} and CLIP (\textbf{C}ontrastive \textbf{L}anguage-\textbf{I}mage \textbf{P}re-training)\cite{IEEEhowto:Radford2021}. CLIP is instead focused on learning visual models from scratch via natural language supervision and does not densely connect the two domains with a joint attention model. Any language tag input can be used as a tag for image classification and regression analysis.

\begin{table*}
\caption{Aesthetic attribute keywords and frequency of PCCD dataset. We use top 5 keywords in DPC-Captions version 1, and top 1000 keywords for DPC-CaptionsV2. In DPC-CaptionsV2, we combine the depth of field and the focus and the composition}
\label{tb:keywords}
\centering
\begin{tabular}{ccc}
\hline
\textbf{Aesthetics Attributes}                    & Aesthetics Attributes in DPC-Captionv2 & Keywords (Frequency)                                                                     \\ \hline
\multirow{2}{*}{Color Lighting} & Light                            & light (1708), sky (493), shadows (491),...                                                  \\ \cline{2-3} 
                                                  & Color (11045)                           & \multicolumn{1}{l}{color (5637), black\&white (3402), blue (1120), red (1097),...}           \\ \hline
Composition                 & \multirow{3}{*}{Composition (8302)}     & \multirow{3}{*}{field (5822), left (2691), perspective (1787), shot (1715), lines (1369),...} \\ \cline{1-1}
Depth of Field                     &                                        &                                                                                          \\ \cline{1-1}
Focus                              &                                        &                                                                                          \\ \hline
General Impression            & Deleted                                & general (4357), good (1810), great (1338), nice (1040),...                                       \\ \hline
Subject of Photo                 & Subject (9331)                          & interesting (708), beautiful (386), light (209), capture (200),...                               \\ \hline
Use of Camera                   & Deleted                                & speed (1488), shutter (1113), iso (1049), aperture (665),...                                     \\ \hline
\end{tabular}
\end{table*}

\begin{figure*}
	\centering
	\includegraphics[height=9.5cm]{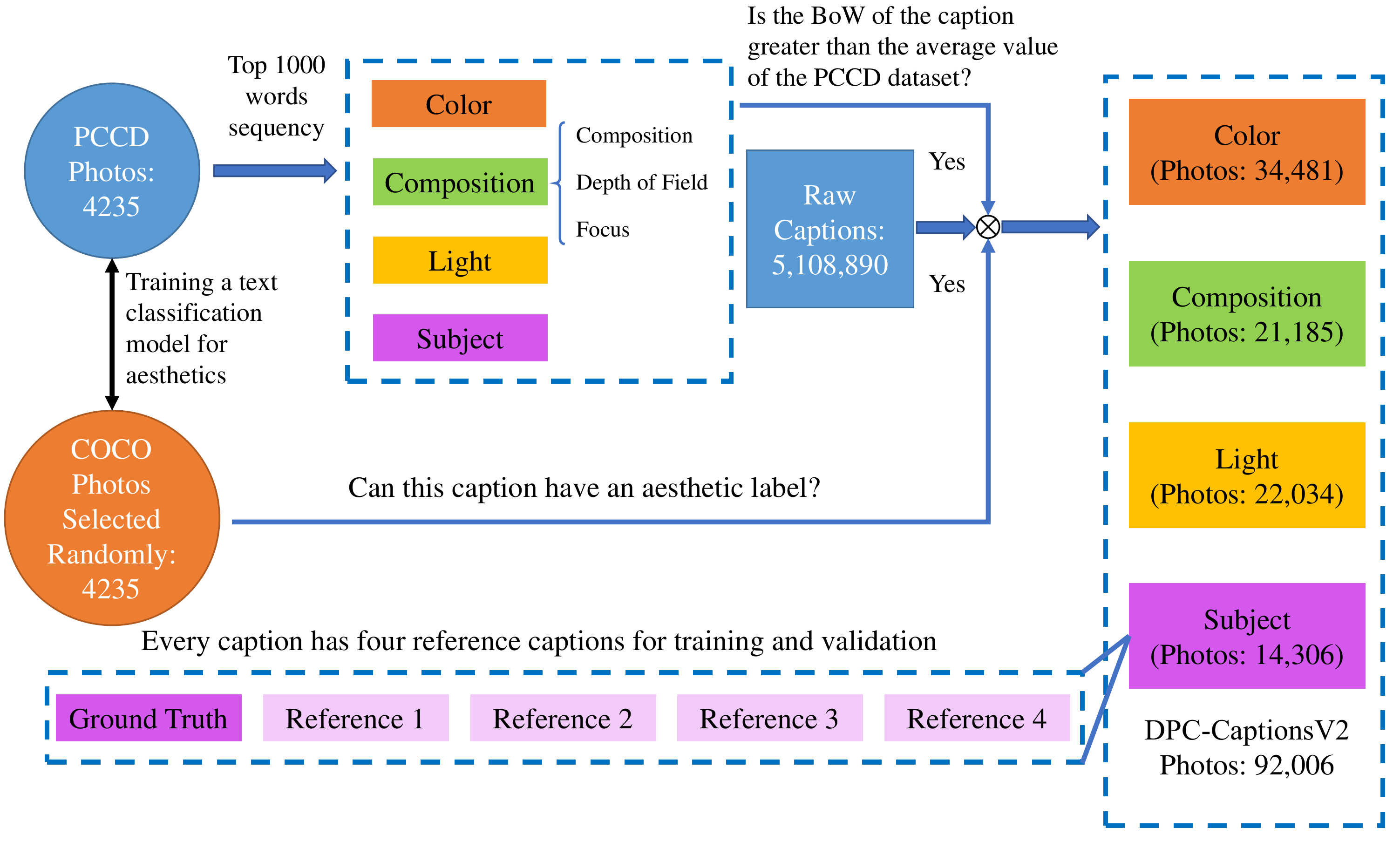}
	\caption{The knowledge transfer method from PCCD to our DPC-CaptionsV2. The PCCD dataset includes 4 aesthetic attributes such as Color Lighting, Composition. The 1000 keywords with the highest word frequency are selected from the comments of each aesthetic attribute. Meantime, each caption need have an aesthetic classify label. When a keyword appears in the comments of an image from DPC-Captions dataset, the image will be assigned to the corresponding aesthetic attribute. The repeated keywords make images be assigned into multiple attributes.}
	\label{fig:dataset}
\end{figure*}

\begin{table*}
	\caption{Comparison of different datasets. The average represents the number of  comments divided by the number of images.}
	\label{tb:dataset} 
	\centering
	\begin{tabular}{llllll}
		\hline\noalign{\smallskip}
		Dataset & Number of Images & Number of Comments & Average Captions & With Attributes & Bias of BoW for Each Caption\\
		\noalign{\smallskip}\hline\noalign{\smallskip}
		AVA-Reviews \cite{IEEEhowto:AVAReviews} & 40,000 & 240,000 & 6 & No & 2,305\\
		AVA-Comments \cite{IEEEhowto:AVAComments}  & \textbf{255,530} & 1,535,937 & 6 & No & 3,991\\
                PCCD \cite{IEEEhowto:ChangICCV17} & 4,235 & 29,645 & 7 & \textbf{Yes} & 7,456\\
		DPC-Captions \cite{IEEEhowto:Jin_19_ACMMM} & 154,384 & \textbf{2,427,483} & 5 & \textbf{Yes} & 6,003\\
		\textbf{DPC-CaptionsV2} & 92,006 & 395,625 & \textbf{11} & \textbf{Yes} & \textbf{8,105}\\
		\noalign{\smallskip}\hline
	\end{tabular}
\end{table*}

\section{DPC-CaptionsV2 via Knowledge Transfer}
During the data cleaning process of the DPC-Caption version 1, we get help from PCCD dataset \cite{IEEEhowto:ChangICCV17}. The aesthetic attributes of PCCD dataset include \emph{Color Lighting}, \emph{Composition}, \emph{Depth of Field}, \emph{Focus}, \emph{General Impression} and \emph{Use of Camera}. For each aesthetic attribute, the five most frequent keywords are selected from the captions. We omit the adverbs, prepositions and conjunctions. We merge words of similar meaning such as color and colour, color and colors. A statistic of the keywords frequency is shown in Table \ref{tb:keywords}.  

Counting the frequency of few words causes that the definition of image aesthetics is limited by a few words and lacking diversity. Due to using too many words in the selecting process, models will be required to adjust the weight of a few high-frequency words in order to filter out real and effective comments.

In our attempts, the text classification and clustering model will not be able to separate the adjectives describing the aesthetics of the image from the specific objects. For example, both \emph{blue} and \emph{sky} appear once in \emph{blue sky}, but only the color description word \emph{blue} is related to the aesthetic of the image, instead of the the word \emph{sky}.

We use Bow (Bag of Words) and a BERT-based text classification model to filter out the necessary comments. Among them, Bow uses the 1,000 most frequently occurring words or phrases in the PCCD data set, and removes some nouns and stopwords. We count the vocabulary word frequency in each sentence as the aesthetic weight and rank, and take the top 100,000 comments. In the BERT-based text binary classification model, we use all PCCD comments as positive samples, and randomly select the same number of 29,645 comments from the COCO data set as negative samples. We judge whether a sentence meets the aesthetic attributes. After satisfying the Bow weight ranking, it still has an aesthetic classification result.

Compared with version 1, DPC-CaptionsV2 has the more keywords for the classification of aesthetic captions, such as \emph{composition}, \emph{color}, \emph{light}, and \emph{subject}. as tags. The tags of captions we used contain more vocabulary related to aesthetic attributes. After two screening weeks, there are still 92,006 DPC subtitle images. In the absence of image-related photography attributes or \emph{use of camera}, it is difficult for the subtitles to learn relevant features directly from the image. So  we deleted the image related to the \emph{use of camera} attribute. We also found that the keywords, \emph{composition}, \emph{depth of field} and \emph{focus} are similar, so we merged them. Finally, we obtained the 4 attributes of DPC-CaptionsV2, as shown in the table \ref{tb:keywords}. The number of images with \emph{light} is 22,034. And, the number of images with \emph{subject} is 14,306, and so on.

In subsequent experiments, the experiment shows that the task of image captioning in the aesthetics is more difficult to learn and generate than the task of general image captioning. Based on this, we adjust the captions’ length in the DPC-CaptionsV2 dataset to make it shorter than DPC-Captions. At the same time, we let each caption has at least 3 reference comments, so we can get easier in training process and validation. Comparing our DPC-CaptionsV2 with the PCCD, AVA-Reviews, AVA-Comments and DPC-Caption data sets in the table \ref{tb:dataset}, it can be seen that AVA-Reviews and AVA-Comments do not contain aesthetics Attributes. Compared with DPC-Captios, DPC-CaptionsV2 provides more authentic and extensive aesthetic captions to improve the quality of the dataset.

\section{Aesthetic Multi-Attribute Network Version 2 (AMANv2)}
\begin{figure*}
	\centering
	\includegraphics[width=18cm]{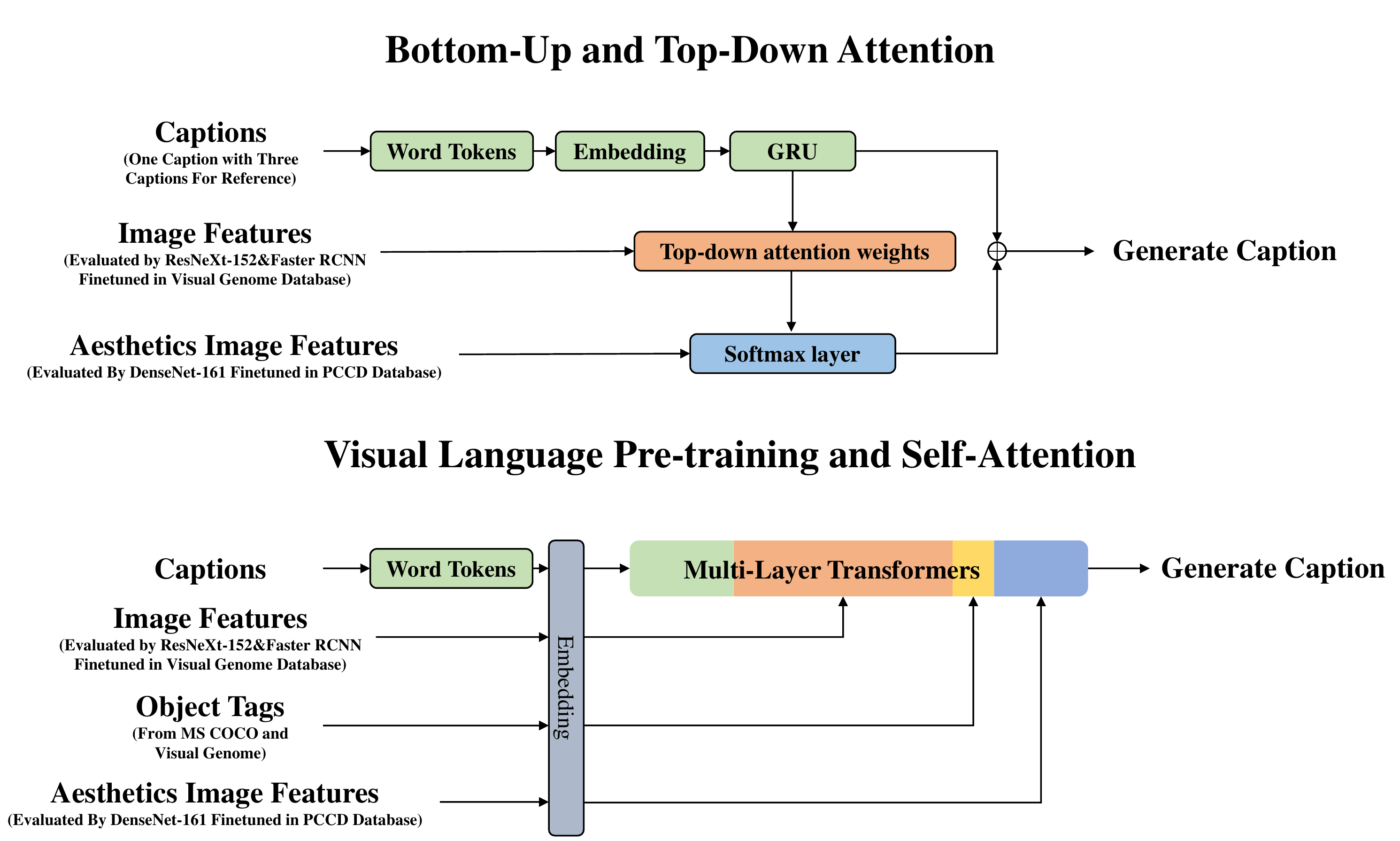}
	\caption{Aesthetic Multi-attribute Network version 2 (AMANv2) includes two model solutions, Bottom-Up and Top-Down Attention (BUTD) and Visual Language Pre-training and Self-Attention (VLPSA). The two models can achieve different performances in different aesthetic attributes. BUTD includes Bottom-Up Attention and Top-Down Attention. Among them, Bottom-Up Attention is Faster R-CNN, which is used to extract the image features of the picture, and the preprocessing model is used to extract the object features and the position information corresponding to the object in the image. Top-Down Attention is used to fuse image features and text features output by GRU units. Our AMANv2 is pre-trained on Visual Genome and fine-tuned on our DPC-Captions data set. The VLPSA model extracts text information, image features, target tags, and image aesthetic features through the transformer unit, and all information is output through multi-layer transfomers to generate captions.}
	\label{fig:network}
\end{figure*}

The PCCD dataset includes captions and attribute scores, while the DPC-Captions dataset only includes captions. In the previous research, for the image in DPC-Captions, only some of the 5 attributes may have annotations attached. We think that PCCD is a fully annotated dataset, and DPC-Captions is a weakly annotated dataset. We recommend learning from a combination of fully annotated datasets and weakly annotated datasets.

Rethinking the process of dataset construction, we find that using a small dataset to guide the construction of larger datasets is improper. It is easy to produce both over-fitting on the results and the limitations of captions. Therefore, we used a larger dataset to guide the generation of DPC-Caption: fine-tuned to generate aesthetic features with a pre-trained model based on ImageNet and Visual Genome.


\begin{figure*}
	\centering
	\includegraphics[height=19cm]{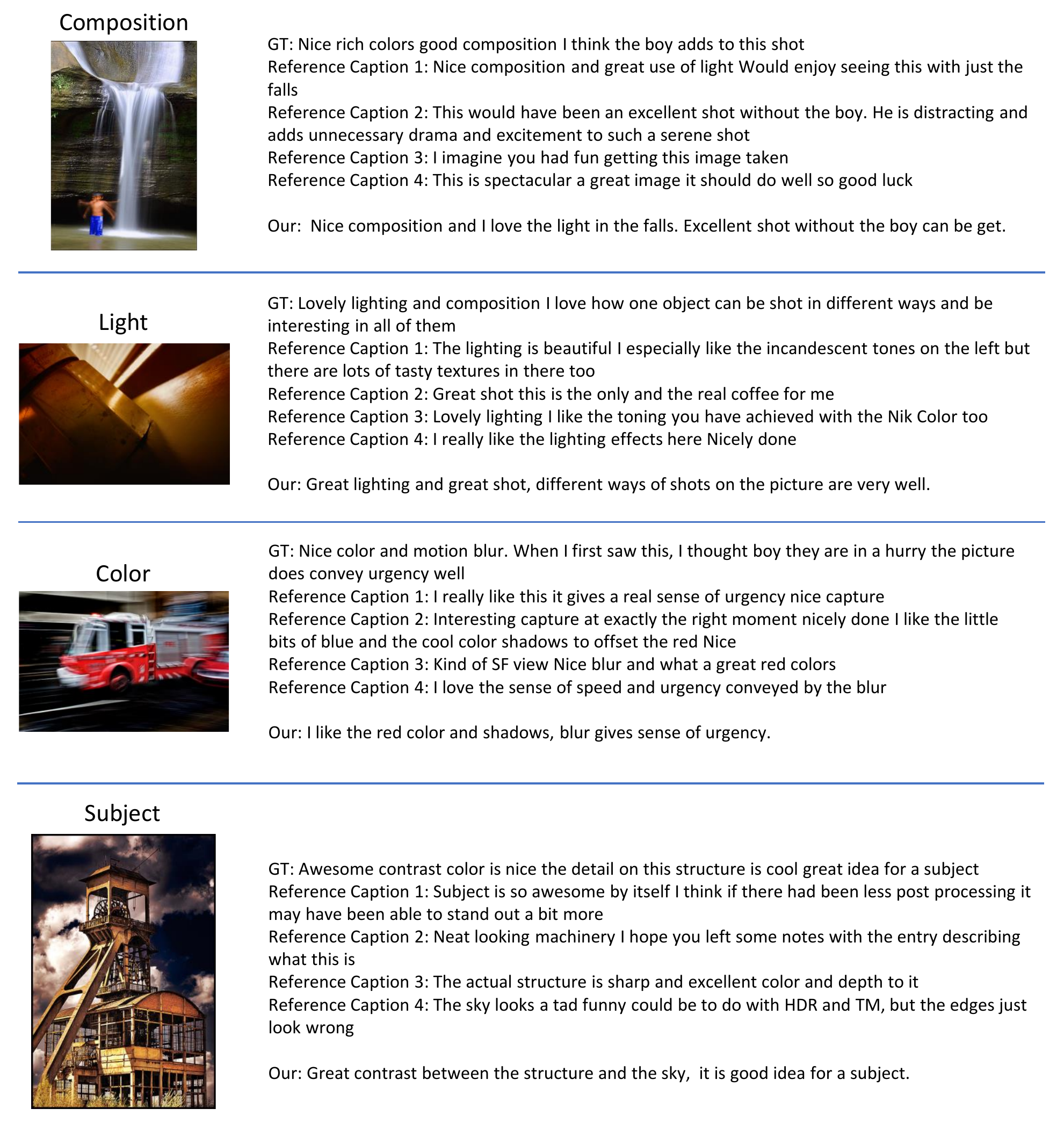}
	\caption{The results of aesthetic multi-attribute network on DPC-Captions dataset. Not all pictures conform to all attribute reviews, so each picture has a Ground Truth and 4 Reference Captions for the prediction of Captions for a certain attribute. The sentences output by the AMAN v2 model synthesize multiple captions, and finally output a comprehensive evaluation of the attribute.}
	\label{fig:avaresult}
\end{figure*}

\subsection{Aesthetic Multi-attribute(AM)}
In the previous task, we used all the data of the PCCD dataset: in addition to the scores for each attribute, there are also global scores for each image. Thus, the loss of multi-attribute aesthetic is divided into two parts. The first one is the loss of each attribute ($m$ attributes, in our paper $m=5$). The second one is the global loss. $N$ represents the number of images in a batch. $\hat{y^{i}}$ represents the output of the last fully connected layer of the network. $y^{i}$ represents the true score. The equal sign in Eq. \ref{eq :lossAG} represents the same calculation method of the global loss and single attribute loss. There are totally 6 loss layers in this model.
	\begin{equation} 
		Loss^{Attribute}= Loss^{Gloal} =\frac{1}{2N}\sum_{i=1}^{N}\left \| \hat{y^{i}} - y^{i} \right \|_{2}^{2}
		\label{eq:lossAG}
	\end{equation}
	\begin{equation}
		Loss=\sum_{j=1}^{m}Loss^{Attribute}_j+Loss^{Gloal}
		\label{eq:loss}
	\end{equation}

The calculation of the above loss function has a basic effect, but ignores that multiple attributes are not completely independent. Therefore, in the training of DPC-CaptionV2, we adjusted the pre-training of the data set into two parts: ImageNet-based pre-training for migration learning to obtain feature vectors of aesthetic significance; Visual Genome-based pre-training to obtain images The feature vector of the target.
	\begin{equation} 
		Loss^{Object} = aesthetic_bias\left \| \hat{y^{i}} - y^{i} \right \|_{2}^{2}
	\end{equation}
	\begin{equation}
		Loss=\sum_{j=1}^{m}Loss^{Attribute}_j+Loss^{Gloal}+Loss^{Object}
		\label{eq:loss}
	\end{equation}

\begin{table*}
	\caption{Performance of the proposed models on the DPC-Captions dataset and DPC-CaptionsV2 dataset. We report BLEU-4 and SPICE. All values refer to percentage ($\%$).}
	\label{tb:performance} 
	\centering
\begin{tabular}{llll}
\hline
\multicolumn{1}{c}{Dataset}                       & Method                            & BLEU4 & SPICE \\ \hline
\multicolumn{1}{c}{\multirow{3}{*}{DPC-Captions}} & CNN-LSTM (Composition)             & 7.0   & 0.167 \\
\multicolumn{1}{c}{}                              & AMAN (Composition)                 & 7.5   & 0.197 \\ \hline
\multirow{3}{*}{DPC-CaptionsV2}                   & CNN-LSTM (Composition)             & 7.4   & 0.192 \\
                                                  & AMANv2-BUTD (Composition)          & 7.73  & \textbf{0.211} \\
                                                  & AMANv2-OSCAR (Composition)         & \textbf{7.74}  & 0.202 \\ \hline
\multicolumn{1}{c}{\multirow{3}{*}{DPC-Captions}} & CNN-LSTM (Color and Lighting)      & 6.9   & 0.166 \\
\multicolumn{1}{c}{}                              & AMAN (Color and Lighting)          & 7.3   & 0.196 \\ \hline
\multirow{6}{*}{DPC-CaptionsV2}                   & CNN-LSTM (Color)                   & 7.21  & 0.181 \\
                                                  & AMANv2-BUTD (Color)                & \textbf{7.69}  & \textbf{0.217} \\
                                                  & AMANv2-OSCAR (Color)               & 7.63  & 0.213 \\ \cline{2-4} 
                                                  & CNN-LSTM (Light)                   & 7.05  & 0.166 \\
                                                  & AMANv2-BUTD (Light)                & \textbf{7.46}  & \textbf{0.190} \\
                                                  & AMANv2-OSCAR (Light)               & 7.44  & 0.188 \\ \hline
\multicolumn{1}{c}{\multirow{3}{*}{DPC-Captions}} & CNN-LSTM (Impression and Subject)  & 6.9   & 0.158 \\
\multicolumn{1}{c}{}                              & AMAN (Impression and Subject)      & 7.4   & 0.181 \\ \hline
\multirow{3}{*}{DPC-CaptionsV2}                   & CNN-LSTM (Subject)                 & 7.15  & 0.184 \\
                                                  & AMANv2-BUTD (Subject)              & \textbf{7.67}  & \textbf{0.204} \\
                                                  & AMANv2-OSCAR (Subject)             & 7.64  & 0.202 \\ \hline
\end{tabular}
\end{table*}

\subsection{Attention Network(AN)}

\textbf{Bottom-Up and Top-Down Attention.} The bottom-up attention model (usually Faster R-CNN) is used to extract the region of interest in the image to obtain object features; the top-down attention model is used to learn the weights corresponding to the features (generally Use LSTM) to achieve in-depth understanding of visual images. Faster R-CNN implements a bottom-up attention model, allowing the overlap of interest frames through a set threshold. So the image content can be understood more effectively. In this paper, not only the object detector but also the attribute classifier are used for each region of interest, so a binary description of the object (attribute, object) can be obtained. The top-down attention model includes a two-layer LSTM model. One is used to implement top-down attention, and the other is used to implement a language model.

\textbf{Visual Language Pre-training and Self-Attention.} This method introduces the label in the target detection result as an anchor point to reduce the learning difficulty of image and text alignment. The model defines each sample (image-text) as a triple (word sequence, object label, regional feature). Use triples as the input of Transformer to generate Captions. Two loss functions are designed at the same time: 1) The mask recovery loss of the text view, which is used to measure the ability of the model to recover the missing elements (words or object tags) according to the context; 2) The contrast loss of the image perspective, which measures the model to distinguish the original The ability of triples and their "contaminated" versions (that is, the original object tags are replaced by randomly sampled tags).

The model uses the semantic information of the image to guide the generation of the word sequence in the decoder stage. Avoiding the problem of using the image information only at the beginning of the decoder leads that the image information is gradually lost with time. In order to obtain the high-level semantic information of the image, the model improves the original convolutional neural networks, including the method of multi-task learning, which can extract the high-level semantic information of the image and enhance the extraction of image features in the encoder stage.

\section{Experiments}

\subsection{Baseline}

\textbf{CNN-LSTM.} This model is based on Goolge's NIC model \cite{IEEEhowto:VinyalsTBE15}. The Resnet-152 \cite{IEEEhowto:HeCVPR2016} extracts features for different attributes and LSTM for encoding. The differences between this baseline and our method include: (1) no attention mechanism is introduced to enhance the feature extraction process; (2) the multi-tasking network is not used to extract features of different attributes. Instead, each attribute trains a network separately. It is not taking full advantage of the aesthetic features, we carry out a simple knowledge transfer in extracting the characteristics of CNNs.

\textbf{AMAN.} Aesthetic Multi-Attribute Network (AMAN) \cite{IEEEhowto:Jin_19_ACMMM} contains Multi-Attribute Feature Network (MAFN), Channel and Spatial Attention Network (CSAN), and Language Generation Network (LGN). The core of MAFN contains GFN and AFN, which regress the global score and attribute scores of an image in PCCD using multi-task regression. They share the dense feature map and have separated global and attribute feature maps, respectively. AMAN is pre-trained on PCCD and finetuned on our DPCCaptions dataset. The CSAN dynamically adjusts the attentional weights of channel dimension and spatial dimension of the extracted features. The LGN generates the final comments by LSTM networks which are fed with ground truth attribute captions in DPC-Captions and attribute feature maps from CSAN.

\subsection{Implementation details}

Our experiments are based on Pytorch framework. The length of LSTM units is 1000. The features of images send to the LSTM unit include ResNet-152 attribute features. The two stage training of AMANv2 is our contribution of using bottom-up and top-down attention and visual language pre-training and self-attention. Except the two stage training, the baseline methods CNN-LSTM use the same training parameters as following:  The word vector dimensionality is set to 300. The underlying learning rate is 0.01. The dimensions of the force module and channel attention module are 512. The dropout is used in training to prevent overfitting. The network is optimized using a stochastic gradient descent optimization strategy. The batch size is set to 64 for DPC-CaptionsV2 and 16 for PCCD. 

\subsection{Attribute Captioning Results}
We train and test our methods on the DPC-Captions and PCCD datasets. Some test results on the DPC-Captions dataset are shown in Figure \ref{fig:avaresult} . It is worth noting that the results are not only rich in sentence structure but also very accurate in grasping features. The relevance of comments and attributes are high. Our results can produce a variety of attribute results. The PCCD author's method \cite{IEEEhowto:ChangICCV17} can only produce one sentence. In addition, our results tend to be objectively evaluated, and the PCCD author's approach favors subjective evaluation.

\subsection{Comparisons}
\label{sec:comparisions}

The evaluation criteria to compare the performance of our model and the baseline models include BLEU-4, which are commonly used in nature language processing community. 

We use SPICE \cite{IEEEhowto:SPICEECCV16} to compare the performance between previous methods and our model. SPICE is a criteria for the automatic evaluation of generated image captions. It resolves the similarity between the result and the generated captions by parsing the sentence into a graph. The calculation formula is as follows.

\begin{footnotesize}
	\begin{equation}
		SPICE=F_{1}Score=\frac{2*Precision*Recall}{precision+Recall}
	\end{equation}
\end{footnotesize}

As shown in Table \ref{tb:performance}, the comparison results shown in Table \ref{tb:performance} reveal that our model outperforms the previous models. Both the BUTD and the OSCAR based AMANv2 models achieve better BLEU4 and SPICE on our new DPC-CaptionsV2.


\section{Conclusion and Discussion}
In this paper, we propose a new task of IAQA: aesthetic attributes assessment. A new vision-language dataset called DPC-Captions is built by knowledge transfer for this task. We propose a novel network AMANv2 for two-stage learning processes on both full annotated small-scale dataset and weakly annotated large-scale dataset, which based on Bottom-Up and Top-Down Attention (BUTD) and Visual Language Pre-training and Self-Attention (VLPSA). Our AMANv2 can generate captions of individual aesthetic attributes. 

In the future, we will explore to caption from sentences to paragraphs. The knowledge transfer methods can be used to build larger dataset for weakly supervised learning. The relations among  attributes can be used for caption learning. Reinforcement learning can also be leveraged for captions generation.

\section*{Acknowledgments}
Parts of this work has been appeared in our previous conference version \cite{IEEEhowto:Jin_19_ACMMM}. This work is partially supported by the National Natural Science Foundation of China (62072014), the Beijing Natural Science Foundation (L192040), and the CAAI-Huawei MindSpore Open Fund (CAAIXSJLJJ-2021-022A).

\ifCLASSOPTIONcaptionsoff
  \newpage
\fi

\bibliographystyle{IEEEtran}

\begin{thebibliography}{}
\providecommand{\url}[1]{#1}
\csname url@samestyle\endcsname
\providecommand{\newblock}{\relax}
\providecommand{\bibinfo}[2]{#2}
\providecommand{\BIBentrySTDinterwordspacing}{\spaceskip=0pt\relax}
\providecommand{\BIBentryALTinterwordstretchfactor}{4}
\providecommand{\BIBentryALTinterwordspacing}{\spaceskip=\fontdimen2\font plus
\BIBentryALTinterwordstretchfactor\fontdimen3\font minus
  \fontdimen4\font\relax}
\providecommand{\BIBforeignlanguage}[2]{{%
\expandafter\ifx\csname l@#1\endcsname\relax
\typeout{** WARNING: IEEEtran.bst: No hyphenation pattern has been}%
\typeout{** loaded for the language `#1'. Using the pattern for}%
\typeout{** the default language instead.}%
\else
\language=\csname l@#1\endcsname
\fi
#2}}
\providecommand{\BIBdecl}{\relax}
\BIBdecl

\end{thebibliography}


\begin{thebibliography}{1}
    \bibitem{IEEEhowto:Jin_19_ACMMM}
    Jin, Xin, et al. "Aesthetic attributes assessment of images." Proceedings of the 27th ACM International Conference on Multimedia. 2019.
	
	\bibitem{IEEEhowto:SPICEECCV16}
	Anderson, Peter, et al. "Spice: Semantic propositional image caption evaluation." European conference on computer vision. Springer, Cham, 2016.
	
	\bibitem{IEEEhowto:Anderson_2018_CVPR}
	Anderson, Peter, et al. "Bottom-up and top-down attention for image captioning and visual question answering." Proceedings of the IEEE conference on computer vision and pattern recognition. 2018.
	
	\bibitem{IEEEhowto:Aneja_2018_CVPR}
	Aneja, Jyoti, Aditya Deshpande, and Alexander G. Schwing. "Convolutional image captioning." Proceedings of the IEEE conference on computer vision and pattern recognition. 2018.
	
    \bibitem{IEEEhowto:ZhouACMMM16}
   Zhou, Ye, et al. "Joint image and text representation for aesthetics analysis." Proceedings of the 24th ACM international conference on Multimedia. 2016.
	
    \bibitem{IEEEhowto:ChangICCV17}
	Chang, Kuang-Yu, Kung-Hung Lu, and Chu-Song Chen. "Aesthetic critiques generation for photos." Proceedings of the IEEE International Conference on Computer Vision. 2017.
	
	\bibitem{IEEEhowto:Chen_2018_CVPR}
	Chen, Fuhai, et al. "Groupcap: Group-based image captioning with structured relevance and diversity constraints." Proceedings of the IEEE conference on computer vision and pattern recognition. 2018.
	
	\bibitem{IEEEhowto:SCACNNCVPR17}
	Chen, Long, et al. "Sca-cnn: Spatial and channel-wise attention in convolutional networks for image captioning." Proceedings of the IEEE conference on computer vision and pattern recognition. 2017.
	
	\bibitem{IEEEhowto:ChenTIP2015}
	Chen, Xiaowu, et al. "Learning templates for artistic portrait lighting analysis." IEEE transactions on image processing 24.2 (2014): 608-618.
	
	\bibitem{IEEEhowto:CuiTMM2018}
	Cui, Chaoran, et al. "Distribution-oriented aesthetics assessment with semantic-aware hybrid network." IEEE Transactions on Multimedia 21.5 (2018): 1209-1220.
	
	\bibitem{IEEEhowto: DengSPM2017}
	Deng, Yubin, Chen Change Loy, and Xiaoou Tang. "Image aesthetic assessment: An experimental survey." IEEE Signal Processing Magazine 34.4 (2017): 80-106.
	
	\bibitem{IEEEhowto: DonahueHRVGSD17}
	Donahue, Jeffrey, et al. "Long-term recurrent convolutional networks for visual recognition and description." Proceedings of the IEEE conference on computer vision and pattern recognition. 2015.
	
	\bibitem{IEEEhowto: DongNC2015}
	Dong, Zhe, and Xinmei Tian. "Multi-level photo quality assessment with multi-view features." Neurocomputing 168 (2015): 308-319.
	
	\bibitem{IEEEhowto:HeCVPR2016}
	He, Kaiming, et al. "Deep residual learning for image recognition." Proceedings of the IEEE conference on computer vision and pattern recognition. 2016.
	
	\bibitem{IEEEhowto:HuangLMW17}
	Huang, Gao, et al. "Densely connected convolutional networks." Proceedings of the IEEE conference on computer vision and pattern recognition. 2017.
	
	\bibitem{IEEEhowto:JinWCSP2016}
	Jin, Xin, et al. "Deep image aesthetics classification using inception modules and fine-tuning connected layer." 2016 8th International Conference on Wireless Communications and Signal Processing (WCSP). IEEE, 2016.
	
	\bibitem{IEEEhowto:JinAAAI2018}
	Jin, Xin, et al. "Predicting aesthetic score distribution through cumulative jensen-shannon divergence." Proceedings of the AAAI Conference on Artificial Intelligence. Vol. 32. No. 1. 2018.
	
	\bibitem{IEEEhowto:JinECCV2010}
	Jin, Xin, et al. "Learning artistic lighting template from portrait photographs." European conference on Computer vision. Springer, Berlin, Heidelberg, 2010.
	
	\bibitem{IEEEhowto:KaoArXiv2016}
	Kao, Yueying, Ran He, and Kaiqi Huang. "Deep aesthetic quality assessment with semantic information." IEEE Transactions on Image Processing 26.3 (2017): 1482-1495.
	
	\bibitem{IEEEhowto:KaoSPIC2016}
	Kao, Yueying, Kaiqi Huang, and Steve Maybank. "Hierarchical aesthetic quality assessment using deep convolutional neural networks." Signal Processing: Image Communication 47 (2016): 500-510.
	
	\bibitem{IEEEhowto:KarpathyF17}
	Karpathy, Andrej, and Li Fei-Fei. "Deep visual-semantic alignments for generating image descriptions." Proceedings of the IEEE conference on computer vision and pattern recognition. 2015.
	
	\bibitem{IEEEhowto:KongECCV2016}
	Kong, Shu, et al. "Photo aesthetics ranking network with attributes and content adaptation." European Conference on Computer Vision. Springer, Cham, 2016.
	
	\bibitem{IEEEhowto:LuMM2014}
	Lu, Xin, et al. "Rapid: Rating pictorial aesthetics using deep learning." Proceedings of the 22nd ACM international conference on Multimedia. 2014.
	
	\bibitem{IEEEhowto:Luo_2018_CVPR}
	Luo, Ruotian, et al. "Discriminability objective for training descriptive captions." Proceedings of the IEEE Conference on Computer Vision and Pattern Recognition. 2018.
	
	\bibitem{IEEEhowto:MaCVPR2017}
	Ma, Shuang, Jing Liu, and Chang Wen Chen. "A-lamp: Adaptive layout-aware multi-patch deep convolutional neural network for photo aesthetic assessment." Proceedings of the IEEE Conference on Computer Vision and Pattern Recognition. 2017.
	
	\bibitem{IEEEhowto:MaiCVPR2016}
	Mai, Long, Hailin Jin, and Feng Liu. "Composition-preserving deep photo aesthetics assessment." Proceedings of the IEEE conference on computer vision and pattern recognition. 2016.
	
	\bibitem{IEEEhowto:MaoHTCY016}
	Mao, Junhua, et al. "Generation and comprehension of unambiguous object descriptions." Proceedings of the IEEE conference on computer vision and pattern recognition. 2016.
	
	\bibitem{IEEEhowto:Mathews_2018_CVPR}
	Mathews, Alexander, Lexing Xie, and Xuming He. "Semstyle: Learning to generate stylised image captions using unaligned text." Proceedings of the IEEE Conference on Computer Vision and Pattern Recognition. 2018.
	
	\bibitem{IEEEhowto:MurrayCVPR2012}
	Murray, Naila, Luca Marchesotti, and Florent Perronnin. "AVA: A large-scale database for aesthetic visual analysis." 2012 IEEE Conference on Computer Vision and Pattern Recognition. IEEE, 2012.
	
	\bibitem{IEEEhowto:NIMATIP18}
	Talebi, Hossein, and Peyman Milanfar. "NIMA: Neural image assessment." IEEE Transactions on Image Processing 27.8 (2018): 3998-4011.
	
	\bibitem{IEEEhowto:KuangTMM2020}
	Q. Kuang, X. Jin, Q. Zhao and B. Zhou, "Deep Multimodality Learning for UAV Video Aesthetic Quality Assessment," in IEEE Transactions on Multimedia, vol. 22, no. 10, pp. 2623-2634, Oct. 2020.
	
	\bibitem{IEEEhowto:VinyalsTBE15}
	Vinyals, Oriol, et al. "Show and tell: A neural image caption generator." Proceedings of the IEEE conference on computer vision and pattern recognition. 2015.
	
	\bibitem{IEEEhowto:AVAReviews}
	Wang, Wenshan, et al. "Neural aesthetic image reviewer." IET Computer Vision 13.8 (2019): 749-758.
	
	\bibitem{IEEEhowto:WangSP2016}
	Wang, Weining, et al. "A multi-scene deep learning model for image aesthetic evaluation." Signal Processing: Image Communication 47 (2016): 511-518.
	
	\bibitem{IEEEhowto:AVAComments}
	Zhou, Ye, et al. "Joint image and text representation for aesthetics analysis." Proceedings of the 24th ACM international conference on Multimedia. 2016.

	\bibitem{IEEEhowto:Pfister2021}
	Pfister, Jan, Konstantin Kobs, and Andreas Hotho. "Self-Supervised Multi-Task Pretraining Improves Image Aesthetic Assessment." Proceedings of the IEEE/CVF Conference on Computer Vision and Pattern Recognition. 2021.
	
	\bibitem{IEEEhowto:Liu2019}
	Liu, Dong, et al. "Modeling image composition for visual aesthetic assessment." Proceedings of the IEEE/CVF Conference on Computer Vision and Pattern Recognition Workshops. 2019.

	\bibitem{IEEEhowto:Zeng2019}
	Zeng, Hui, et al. "A unified probabilistic formulation of image aesthetic assessment." IEEE Transactions on Image Processing 29 (2019): 1548-1561.

	\bibitem{IEEEhowto:Lu2019}
	Lu, Jiasen, et al. "ViLBERT: pretraining task-agnostic visiolinguistic representations for vision-and-language tasks." Proceedings of the 33rd International Conference on Neural Information Processing Systems. 2019.

	\bibitem{IEEEhowto:Tan2019}
	Tan, Hao, and Mohit Bansal. "LXMERT: Learning Cross-Modality Encoder Representations from Transformers." Proceedings of the 2019 Conference on Empirical Methods in Natural Language Processing and the 9th International Joint Conference on Natural Language Processing (EMNLP-IJCNLP). 2019.
	
	\bibitem{IEEEhowto:Chen2020}
	Chen, Yen-Chun, et al. "Uniter: Universal image-text representation learning." European conference on computer vision. Springer, Cham, 2020.

	\bibitem{IEEEhowto:Su2019}
	Su, Weijie, et al. "VL-BERT: Pre-training of Generic Visual-Linguistic Representations." International Conference on Learning Representations. 2019.

	\bibitem{IEEEhowto:Li2020}
	Li, Xiujun, et al. "Oscar: Object-semantics aligned pre-training for vision-language tasks." European Conference on Computer Vision. Springer, Cham, 2020.

	\bibitem{IEEEhowto:Radford2021}
	Radford, Alec, et al. "Learning transferable visual models from natural language supervision." arXiv preprint arXiv:2103.00020 (2021).

	\bibitem{IEEEhowto:Ghosal2019}
	Ghosal, Koustav, Aakanksha Rana, and Aljosa Smolic. "Aesthetic image captioning from weakly-labelled photographs." Proceedings of the IEEE/CVF International Conference on Computer Vision Workshops. 2019.

	\bibitem{IEEEhowto:Zou2020}
	Zou, Xiaohan, et al. "To be an Artist: Automatic Generation on Food Image Aesthetic Captioning." 2020 IEEE 32nd International Conference on Tools with Artificial Intelligence (ICTAI). IEEE, 2020.
	
	
\end{thebibliography}

%
\begin{IEEEbiography}[{\includegraphics[width=1in,height=1.25in,clip,keepaspectratio]{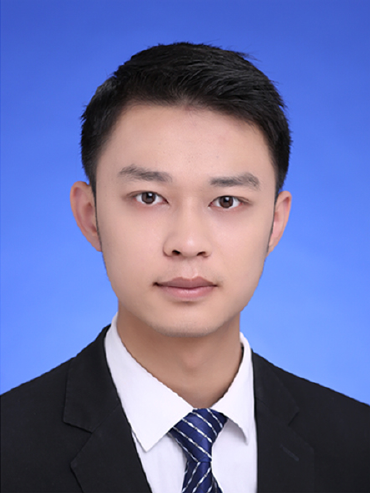}}]{Xinghui Zhou}
is presently a Ph.D. candidate in the University of Science and Technology of China. His research interests are computer vision and image processing.
\end{IEEEbiography}

\begin{IEEEbiography}[{\includegraphics[width=1in,height=1.25in,clip,keepaspectratio]{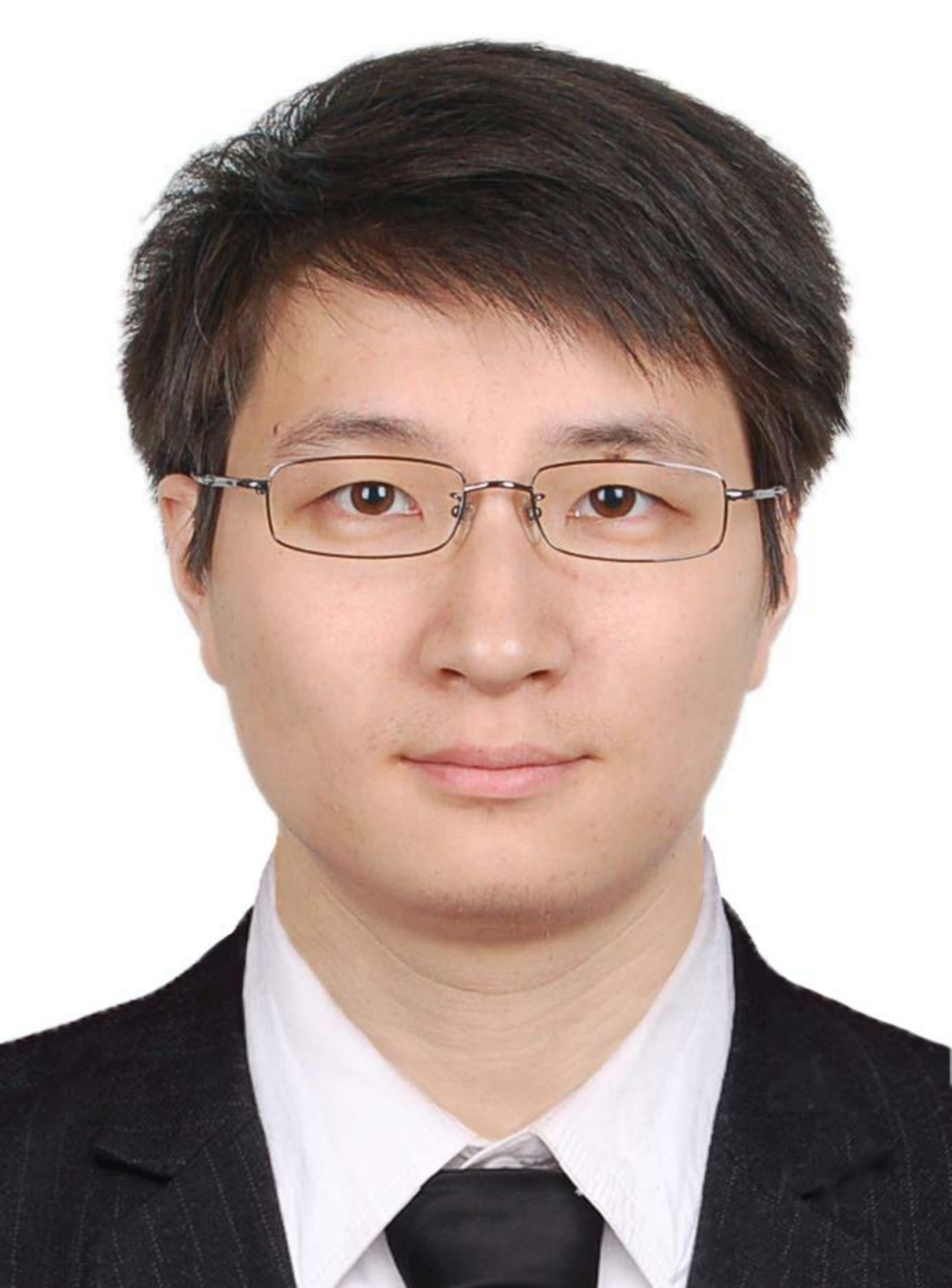}}]{Xin Jin}
is currently an Associate Professor with the Department of Cyber Security at Beijing Electronic and Science Technology Institute and a Visiting Scholar at Beijing Institute for General Artificial Intelligence (BigAI). He received his Ph.D. degrees in Computer Science from Beihang University, Beijing China, in 2013. He was a visiting student at Lotus Hill Institute, Ezhou, China and a visiting scholar at Tsinghua University, Beijing China. His research interests include computational aesthetics, computer vision and artificial intelligence security. He is an associate editor of Cognitive Robotics. He served as the program committee members and session chairs of multiple conferences.
\end{IEEEbiography}

%
%
%
%
%

\begin{IEEEbiography}[{\includegraphics[width=1in,height=1.25in,clip,keepaspectratio]{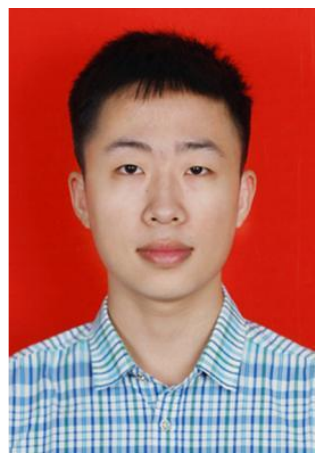}}]{Jianwen Lv}
is currently studying for a master's degree in the Beijing Electronic Science and Technology Institute, Visual Computing And Information Security Lab. His research interests are computer vision and natural language processing. 
\end{IEEEbiography}

\begin{IEEEbiography}[{\includegraphics[width=1in,height=1.25in,clip,keepaspectratio]{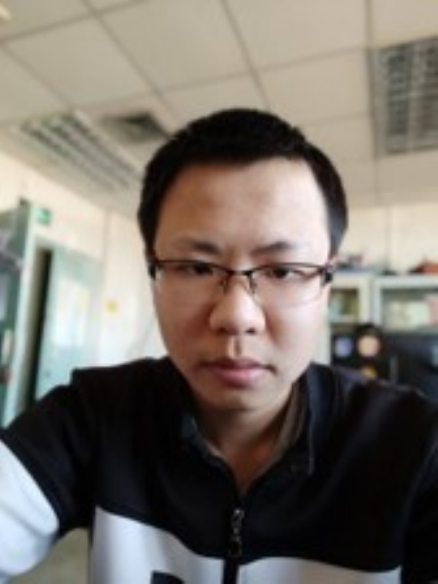}}]{Heng Huang}
is a master student majoring in Cyberspace Security at Beijing Electronic Science and Technology Institute. His research interests are computer vision and artificial intelligence.
\end{IEEEbiography}

\begin{IEEEbiography}[{\includegraphics[width=1in,height=1.25in,clip,keepaspectratio]{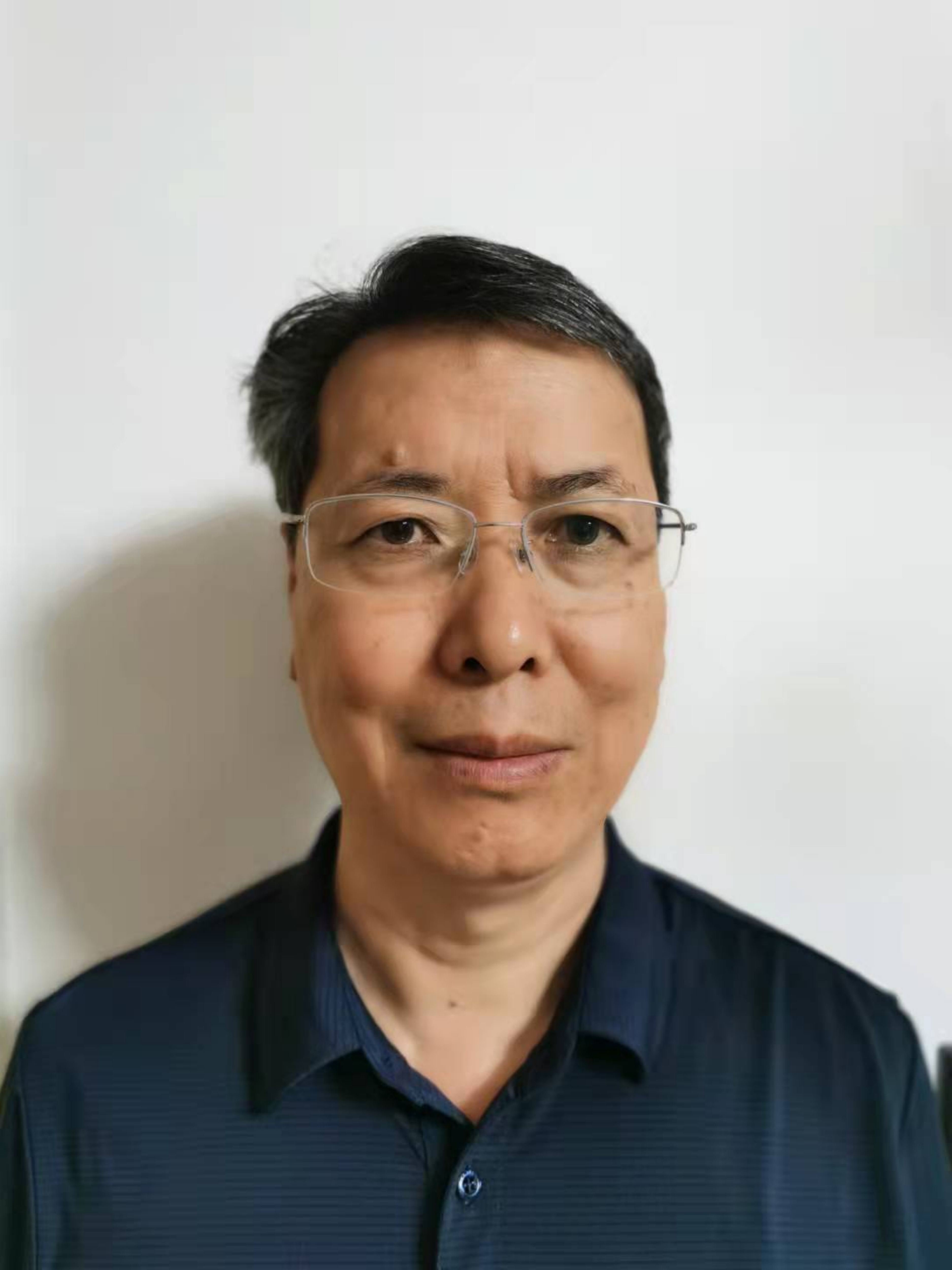}}]{Ming Mao}
is a professor at Beijing Electronic Science and Technology Institute.
\end{IEEEbiography}

\begin{IEEEbiography}[{\includegraphics[width=1in,height=1.25in,clip,keepaspectratio]{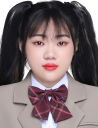}}]{Shuai Cui}
received Bachelor of Arts and Sciences degree in philosophy and mathematics from University of California, Davis, CA, USA in 2021. She was a physics student until the junior year. The Symbolic Logic course open the philosophy door for her. Now, she is applying master programs. Her research interests include logic, metaphysics, epistemology, and Artificial Intelligence.
\end{IEEEbiography}





\end{document}